# AI-Generated Text Detection in Low-Resource Languages: A Case Study on Urdu

1st Muhammad Ammar
*Department of Computer Science*
*COMSATS University Islamabad*
Islamabad, Pakistan
ammarshafique677@gmail.com

2nd Hadiya Murad Hadi
*Department of Computer Science*
*COMSATS University Islamabad*
Islamabad, Pakistan
hadiyamurad96@gmail.com

3rd Usman Majeed Butt
*Department of Computer Science*
*COMSATS University Islamabad*
Islamabad, Pakistan
u4usmanmajeed313@gmail.com

*Abstract*— The growing capabilities of Large Language Models (LLMs) have significantly improved automated text generation, but they have also introduced challenges in distinguishing machine-generated content from human-written text. This issue is especially critical for low resource languages such as Urdu, where effective AI detection tools remain scarce. To address this gap, we propose a novel AI-generated text detection framework tailored for the Urdu language. A balanced dataset comprising 1,800 humans authored, and 1,800 AI generated texts, sourced from models such as Gemini, GPT-4o-mini, and Kimi AI was developed. Detailed linguistic and statistical analysis was conducted, focusing on features such as character and word counts, vocabulary richness (Type Token Ratio), and N-gram patterns, with significance evaluated through T-tests and Mann-Whitney U tests. Three state-of-the-art multilingual transformer models such as Microsoft/mdeberta-v3-base, distillbert/distilbert-base-multilingual-cased, and Facebook/xlm-roberta-base were fine-tuned on this dataset. The mDeBERTa-v3-base achieved the highest performance, with an F1-score 91.29 and accuracy of 91.26% on the test set. This research advances efforts in combating misinformation and academic misconduct in Urdu-speaking communities, and contributes to the broader development of NLP tools for underrepresented languages.

## I. Introduction

The rapid advancement of Large Language Models (LLMs) has significantly transformed the field of natural language processing, enabling machines to generate human-like text with remarkable fluency and coherence. Models such as GPT-4, Gemini, and other cutting-edge LLMs are now capable of producing essays, articles, and conversational responses that are often indistinguishable from content written by humans [1], [2]. While these developments offer substantial benefits in areas such as education, content creation, and automation, they also introduce new and pressing challenges, particularly the ability to reliably detect whether a piece of text has been authored by a human or generated by an AI system.

This issue becomes even more critical in the context of low-resource languages like Urdu. Unlike high-resource languages such as English, which benefit from abundant datasets and tools, Urdu lacks robust AI detection systems and large-scale annotated corpora [5], [6], [7]. As a result, Urdu-speaking communities are especially vulnerable to the misuse of generative AI technologies. The potential consequences include academic dishonesty, where students may submit AI-generated assignments, and the proliferation of misinformation or fake news, which can erode public trust and impact social and political discourse.

Despite the growing adoption of generative AI technologies, there is a significant research gap in the detection of AI-generated content in Urdu. Most existing detectors are trained on English datasets or earlier-generation AI outputs, making them ill-equipped to handle the sophisticated text produced by modern LLMs in diverse languages.

To address this critical need, this study proposes a robust and scalable AI-generated text detection system specifically tailored for the Urdu language. We introduce a custom, balanced dataset comprising human-written and AI-generated Urdu texts and evaluate the performance of state-of-the-art multilingual transformer models fine-tuned for this task. By bridging the gap in AI detection tools for Urdu, this research aims to support efforts in preserving academic integrity and mitigating the spread of AI-driven misinformation in underrepresented language communities.

## II. Related Work

### A. General AI-Generated Text Detection Methods

The field of AI-generated text detection has advanced significantly, transitioning from early rule-based and heuristic approaches to more sophisticated machine learning and deep learning methodologies. Contemporary detection strategies can be broadly categorized into four key paradigms.

One prevalent method is AI model-based detection, where pre-trained models such as RoBERTa are fine-tuned to distinguish between human-written and machine-generated content. However, these approaches often face the "moving target" challenge [1], [2], [3]. As generative models evolve rapidly, detectors trained on earlier LLMs (e.g., RoBERTa trained on GPT-2 outputs) struggle to accurately classify text from newer models like GPT-4, resulting in diminished performance over time.

The second category includes perplexity- and perturbation-based detection. In these approaches, perplexity measures the predictability of a text sequence [2], [14], AI-generated text often has lower perplexity due to its fluency and regularity. Perturbation methods further assess a text's resilience to changes such as sentence shuffling, word substitutions, or paraphrasing. Human-written texts generally retain coherence under minor modifications, while AI-generated content tends to break down more noticeably, revealing its rigid structure. Techniques like GPT Completion Perturbation, sentence insertion, synonym substitution, and paraphrasing are commonly used to test this brittleness.

Another strategy combines Sparse Autoencoders (SAE) with classifiers like XGBoost [3], [15]. SAE models are employed to extract compressed, unsupervised representations from text embeddings, capturing hidden patterns typical of AI-generated writing. These are then fed into XGBoost for classification. While this hybrid method is model-agnostic and powerful, it typically requires large volumes of training data.

Lastly, AI watermarking techniques aim to embed detectable patterns into generated text. These include:

- Probabilistic watermarking, which adjusts token probability distributions.
- Cryptographic watermarking, which encodes hidden signatures using whitespace or embedding shifts.
- Style-based watermarking, which enforces stylistic fingerprints tracked by classifiers.
- Invisible character watermarking, which inserts zero-width Unicode characters.
- Semantic watermarking, which subtly alters sentence phrasing.

Although watermarking can be effective [4], it requires cooperation from the text generator and fails entirely when the source model does not include such watermarks.

This research adopts a more robust and adaptable approach by fine-tuning multilingual transformer models on a newly constructed dataset of Urdu texts. These models, pre-trained on diverse linguistic corpora, offer superior generalization and contextual understanding, making them more resilient to the evolving nature of AI-generated content. Moreover, the linguistic and statistical analyses conducted in this study complement deep learning models by identifying quantifiable stylistic and structural differences between AI and human writing, paralleling the goals of perturbation-based methods but with interpretable metrics such as Type-Token Ratio (TTR) and N-gram diversity.

*B. NLP Challenges and Approaches for Low-Resource Languages*

Natural Language Processing (NLP) in low-resource languages like Urdu presents a distinct set of challenges. These languages often lack comprehensive linguistic resources, large, annotated datasets, and standardized NLP tools, which hinders the development of accurate and scalable language models.

Urdu, in particular, poses multiple linguistic and technical hurdles [6], [7]. Its script (Nasta'liq) is complex and less compatible with standard NLP pipelines. Common issues include improper word segmentation, frequent code-mixing with English, and significant morphological variation. The absence of large, labeled datasets further compounds the difficulty of training reliable NLP systems. Additionally, existing tools are typically optimized for high-resource languages and do not generalize well to the stylistic and grammatical diversity of Urdu.

Despite these limitations, several strategies are being employed to enable NLP in low-resource settings. These include:

- Data augmentation to synthetically expand training datasets.
- Crowdsourced data annotation from native speakers.
- Transfer learning from high-resource languages via multilingual pre-trained models.
- Multimodal techniques, which integrate textual data with audio, video, or image data to enrich linguistic understanding.

Recent advances, such as the application of generative models in Named Entity Recognition (NER) [8], [10] for Nepali and hate speech detection across various under-resourced languages, underscore the growing potential of LLMs even in limited-data environments.

This study incorporates several Urdu-specific preprocessing techniques, such as Unicode normalization, diacritic removal, preservation of Urdu punctuation, and elimination of unnecessary whitespace or special characters to ensure optimal input quality. These language-aware strategies address challenges specific to Urdu morphology and writing style, improving model performance while reducing data noise. The removal of diacritics, although it can obscure phonetic nuances, is a practical trade-off to simplify vocabulary and enhance model generalization in a low-resource context.

*C. Existing Research on Text Detection in Urdu*

While research directly targeting AI-generated text detection in Urdu remains limited, related work in fake news detection offers valuable insights. One notable study titled *"Detecting Fake News in Urdu Language Using Machine Learning, Deep Learning, and Large Language Model-Based Approaches"* [8], [9], [10] employed multiple methodologies to classify misinformation in Urdu content.

The researchers utilized two datasets: the Bend the Truth dataset (900 articles) and the Ax-to-Grind dataset (10,083 articles). They implemented extensive preprocessing, including text cleaning, stop word removal, and tokenization via StanfordNLP. For feature extraction, they applied both traditional methods (e.g., TF-IDF, odds ratios) and modern embeddings (e.g., Word2Vec), and trained a range of classifiers, from Naïve Bayes and SVMs to neural networks (CNNs, LSTMs) and fine-tuned transformer models like BERT and GPT-2.

The results demonstrated that transformer models outperformed traditional approaches, with BERT and GPT-2 achieving F1 scores as high as 93% and accuracy up to 95% on the larger dataset. These findings validate the efficacy of pre-trained transformers in capturing deep semantic patterns in Urdu, especially when fine-tuned on sufficiently large and diverse datasets.

## III. DATASET CURATION

To facilitate the detection of AI-generated Urdu text, a custom, balanced, first of its kind, dataset (UHAT Dataset [20]) was meticulously developed, which spans a diverse range of topics, including literature, news, encyclopedic entries to enhance the generalizability of the detection models consisting of 1,800 human-authored texts and 1,800 AI-generated texts, yielding a total of 3,600 original samples. Human written texts were sourced from authentic and reputable Urdu platforms such as Urdu literature collections, BBC Urdu, Urdu Wikipedia and news articles from multiple credible news outlets. These texts represent high-quality, natural Urdu writing styles across different domains. To generate the AI counterpart, each human written text was rephrased independently by three different LLMs, GPT-4o-mini, Gemini, and Kimi AI.

However, due to the input length constraints of transformer architectures, especially when handling longer text sequences, a substantial portion of these documents required pre-processing through sliding window chunking. Specifically, any text exceeding 450 characters was split into overlapping segments while preserving semantic context. This chunking strategy expanded the dataset to 7,667 total chunks, ensuring comprehensive model exposure without discarding valuable long-form content.

A summary of the chunking process is presented in Table I

TABLE I

DATA CHUNKING SUMMARY

| METRIC | Value |
|---|---|
| Original Texts (Total) | 3600 |
| Texts Requiring Chunk | 1657(46%) |
| Total Chunks Created | 7667 |
| Average Chunks per Text | 2.13 |
| Chunked Dataset Shape | (7667, 6) where 6 is feature dimension |
| Label Distribution in Chunks | Human: 4231 AI: 3436 |
| Chunk Length Statistics (Characters) | Min: 45, Max: 450, Avg:412.76, Std: 69.59 |

Following chunking, the dataset was partitioned into training (80%), validation (10%), and test (10%) sets to support model training and evaluation. Care was taken to maintain good label balance across all splits to prevent class imbalance and ensure statistically valid performance comparisons.

A summary of the dataset split distribution is presented in Table II.

TABLE II

DATASET SPLIT DISTRIBUTION

| Dataset Split | Total Samples | Human Samples | AI Samples |
|---|---|---|---|
| Training Set | 6133 | 3384(55.2%) | 2749(44.8%) |
| Validation Set | 767 | 423(55.1%) | 344(44.9%) |
| Test Set | 767 | 424(55.3%) | 343(44.7%) |

This fairly balanced and augmented dataset forms a robust foundation for fine-tuning transformer models. By combining original text diversity with chunk-level segmentation, the curated dataset ensures sufficient coverage of linguistic patterns, improving the model's ability to generalize across unseen samples while retaining a high level of interpretability and balance.

## IV. METHODOLOGY

### A. Linguistic and Statistical Analysis

To understand the distinctions between human and AI-generated Urdu text, a detailed linguistic and statistical analysis was conducted. This involved extracting key features such as text length, word count, sentence count, average word length, punctuation density, character diversity, and sentence length variability.

Vocabulary richness was measured using the Type-Token Ratio (TTR), [14], [15], and N-gram statistics (Bigrams and Trigrams) were analysed for frequency and uniqueness. To validate these differences, T-tests and Mann-Whitney U tests were applied across all metrics.

Statistical tests found clear differences between human and AI-written texts in 5 out of 6 complexity measures. Humans tend to use a wider variety of words, with a higher type-token ratio (0.7091 compared to 0.6741 for AI). On average, human words are slightly longer (3.5293 vs. 3.4809), and their sentence lengths vary much more (13.36 vs. 4.06). AI texts, however, have slightly longer sentences overall (152.17 words compared to 143.74 for humans). When it comes to punctuation use, there was no significant difference between the two ($p = 0.4349$).

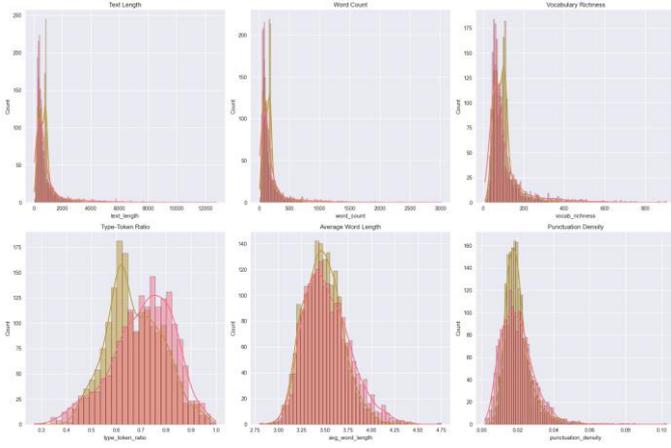

This analysis provides interpretable linguistic features that support the deep learning models. For example, lower TTR or less diverse N-grams in AI texts may indicate repetitive or generic language. Statistical differences in sentence length or punctuation density may reflect the overly structured nature of AI-generated content. These insights help explain what the models are learning.

TABLE III

SUMMARY STATISTICS

| Metric | Human | AI |
|---|---|---|
| Total Texts | 1.800000e+03 | 1.800000e+03 |
| Total Words | 3.562760e+05 | 2.970560e+05 |
| Total Characters | 1.599753e+06 | 1.346400e+06 |
| Unique Words | 1.887700e+04 | 1.162800e+04 |
| Avg Text Length | 8.887517e+02 | 7.480000e+02 |
| Avg Words per Text | 1.979311e+02 | 1.650311e+02 |
| Vocabulary Richness | 1.090183e+02 | 9.636111e+01 |

### B. Urdu Text Preprocessing

Given the linguistic complexity of Urdu, pre-processing was tailored to optimize model performance:

- Unicode Normalization ensured consistent character representation.
- Whitespace and Newline Removal reduced formatting noise.
- Special Character Removal preserved only meaningful Urdu punctuation for sentence boundary clarity.
- Diacritic Removal (Harakat) reduced vocabulary complexity and improved model generalization.

These pre-processing steps reflect a nuanced understanding of Urdu script and were crucial for handling the low-resource nature of the language [6], [7].

### C. Experimental Setup

The dataset was split into 80% training, 10% validation, and 10% testing, ensuring class balance across splits.

To handle input limits of transformer models, sliding window chunking was applied to longer texts (over 450 characters), producing overlapping segments while preserving context. This expanded the dataset significantly and improved training diversity.

Three multilingual transformer models were selected for fine-tuning:

- `microsoft/mdeberta-v3-base`
- `distilbert/distilbert-base-multilingual-cased`
- `FacebookAI/xlm-roberta-base`

These were chosen for their proven multilingual capabilities and suitability for transfer learning in low-resource contexts.

Training Procedures included:

- Early stopping based on validation loss to prevent overfitting.
- Callbacks for model checkpointing and learning rate adjustments.
- AdamW optimizer with tuned learning rates for stable convergence.

This setup ensures robust model performance while maximizing generalization on unseen Urdu text.

### V. RESULTS AND DISCUSSION

#### A. Model Training Summary

The three selected multilingual transformer models (mdeberta-v3-base, distilbert-base-multilingual-cased, FacebookAI/xlm-roberta-base) were fine-tuned on the training set, with performance monitored on the validation set. Early stopping was employed to prevent overfitting and ensure optimal generalization. Table III summarizes the training performance for each model.

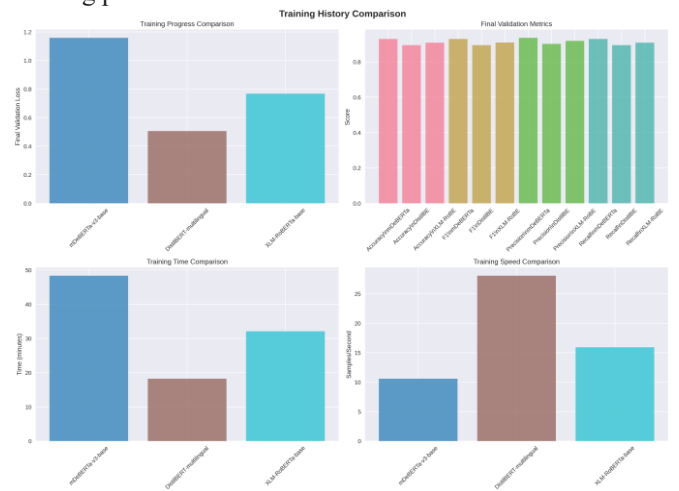

TABLE IV

MODEL TRAINING PERFORMANCE

| Model | F1 Score | Accuracy | Precision |
|---|---|---|---|
| microsoft/mdeberta-v3-base | 92.98% | 92.96% | 93.55% |
| distilbert/distilbert-base-multilingual-cased | 89.47% | 89.44% | 90.11% |
| FacebookAI/xlm-roberta-base | 90.89% | 90.87% | 91.82% |

## B. Analysis of Model Performance

The experimental results validate the effectiveness of multilingual transformer models for AI-generated text detection in Urdu. Among the three fine-tuned models, microsoft/mdeberta-v3-base achieved the best performance, with an F1-score of 92.98%, precision of 93.55%, and accuracy of 92.96% on the training set. Its superior results highlight the benefits of its enhanced architecture and deep multilingual pretraining, which appear well-suited for capturing nuanced language patterns in Urdu.

XLM-RoBERTa-base closely followed, achieving an F1-score of 90.89%, precision of 91.82%, and accuracy of 90.87%, confirming its robustness in multilingual contexts. DistilBERT-multilingual, though slightly behind, still performed competitively, with an F1-score of 89.47%, precision of 90.11%, and accuracy of 89.44%.

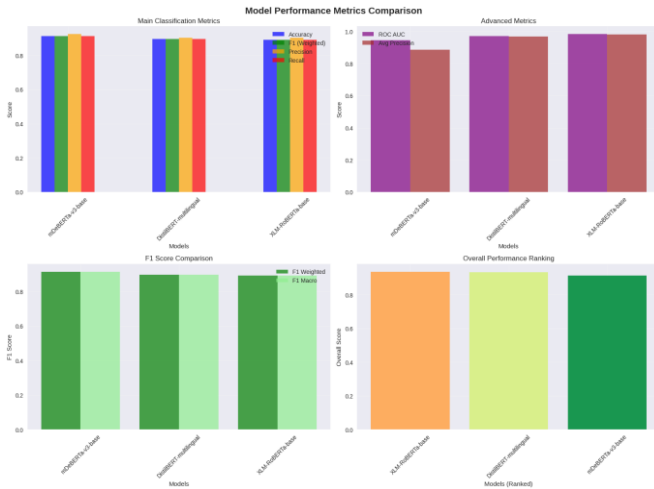

The consistently high scores across all models particularly in both precision and F1-score, demonstrate balanced classification performance for both AI and human-written text. These results also affirm the value of transfer learning in low-resource settings, as the models generalize well to Urdu despite its script complexity and morphological richness. The success of mDeBERTa-v3-base underscores the potential of newer multilingual models to serve as reliable tools for NLP tasks in underrepresented languages [11], [12], [13].

TABLE V

EVALUATION ON TEST SET

| Model | Accuracy | F1 Score | Precision |
|---|---|---|---|
| mDeBERTa-v3-base | 0.9126 | 0.9129 | 0.9232 |
| DistilBERT-multilingual | 0.8957 | 0.8960 | 0.9016 |
| XLM-RoBERTa-base | 0.8905 | 0.8907 | 0.9033 |

## C. Implications for Combating Misuse

The successful development of an AI text detection system for Urdu carries significant real-world implications for mitigating the misuse of generative AI in both educational and digital media contexts.

In academic settings, the system offers a crucial safeguard against academic dishonesty [18], where students may submit AI-generated essays or assignments. The tool empowers educators and institutions to verify the authenticity of student work, reinforcing academic integrity and ensuring fair evaluation standards, an increasingly urgent concern in the age of LLMs.

Beyond education, this system plays a vital role in combating misinformation in Urdu-language media. Generative AI can rapidly produce and propagate fake news, particularly in local languages that lack effective moderation tools. By accurately detecting AI-generated content, this solution equips content moderators, journalists, and fact-checkers to identify and flag misleading or fabricated narratives, helping maintain trust in public discourse [9], [19].

Given the scarcity of existing AI detection tools tailored for Urdu, this research fills a critical gap, transforming theoretical advancements into actionable, deployable technology. It addresses two major societal concerns: preserving academic credibility and curbing the dissemination of misinformation. As such, this work not only contributes to the field of NLP but also offers meaningful, tangible value to Urdu-speaking communities facing the emerging risks of generative AI.

## VI. CONCLUSION

This study successfully addresses the urgent need for AI-generated text detection tools in Urdu, a low-resource language that faces growing challenges with academic dishonesty and AI-driven misinformation. The work involved the creation of a balanced, custom dataset containing human and AI-generated Urdu texts, accompanied by an in-depth linguistic and statistical analysis to extract distinguishing features.

Three state-of-the-art multilingual transformer models were fine-tuned and evaluated. Among them, microsoft/mdeberta-v3-base achieved the best performance, with an F1-score of 91.29%, precision 92.32%, and accuracy of 91.26%. DistilBERT-multilingual followed, with strong results (F1-score: 89.60%, accuracy: 89.57%), while XLM-RoBERTa-base also demonstrated competitive performance (F1-score: 89.07%, accuracy: 89.05%).

These results validate the effectiveness of multilingual transformers for Urdu text classification and demonstrate that reliable AI text detection is achievable even in low-resource languages. This research lays the foundation for practical applications in educational integrity systems, fake news detection, and broader NLP efforts across underrepresented languages.

ACKNOWLEDGEMENT

We extend our appreciation to the developers and maintainers of platforms such as Rekhta.org, BBC Urdu, and Urdu Wikipedia, whose high-quality content played a critical role in building our human-authored dataset. Finally, we recognize the contributions of open-source communities behind the transformer models and NLP libraries used in this research, including Hugging Face and the creators of mDeBERTa, XLM-RoBERTa, and DistilBERT.